\documentclass{article}

\usepackage{arxiv}

\usepackage[utf8]{inputenc} 
\usepackage[T1]{fontenc}    
\usepackage{hyperref}       
\usepackage{url}            
\usepackage{booktabs}       
\usepackage{blindtext}
\usepackage{eso-pic}
\usepackage{nicefrac}       
\usepackage{microtype}      
\usepackage{amsmath,amssymb,amsfonts}
\usepackage{algcompatible}
\usepackage{graphicx}
\usepackage{textcomp}
\usepackage{xcolor}

\usepackage{algorithm}	
\usepackage[noend]{algpseudocode}

\newcommand{\ddx}[1]{\frac{\textrm{d}}{\textrm{d}#1}}
\newcommand{\ddt}{\ddx{t}}
\newcommand{\ppx}[1]{\frac{\partial}{\partial#1}}
\renewcommand{\Re}{\mathbb{R}}   

\title{Continuous Convolutional Neural Networks: Coupled Neural PDE and ODE}

\author{ 
	\href{https://orcid.org/0000-0001-9051-1370}{\includegraphics[scale=0.06]{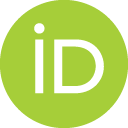}\hspace{1mm}Mansura Habiba} \\
	Dept of Computer Science\\
	Maynooth University\\
	Maynooth, Ireland \\
	\And
	\href{https://orcid.org/0000-0003-0521-4553}{\includegraphics[scale=0.06]{orcid.png}\hspace{1mm}Barak A. Pearlmutter} \\
	Department of Computer Science \& Hamilton Institute\\
	Maynooth University\\
	Maynooth, Ireland \\
}

\begin{document}
\maketitle

\begin{abstract}
	Recent work in deep learning focuses on solving physical systems in the Ordinary Differential Equation or Partial Differential Equation. This current work proposed a variant of Convolutional Neural Networks (CNNs) that can learn the hidden dynamics of a physical system using ordinary differential equation (ODEs) systems (ODEs) and  Partial Differential Equation systems (PDEs). Instead of considering the physical system such as image, time -series as a system of multiple layers, this new technique can model a system in the form of Differential Equation (DEs). The proposed method has been assessed by solving several steady-state PDEs on irregular domains, including heat equations, Navier-Stokes equations. 
\end{abstract}


\section{Introduction}

It is challenging to have an optimal balance between accuracy and efficiency while finding a solution for a Partial Differential Equation (PDE) system. Recent research in deep learning is focused on Physics informed neural networks. At the same time, data-driven neural networks that can be represented as a function of continuous-time series demonstrate the impressive result. Recently, Neural ODE (NODE) \cite{chen2018neural} is becoming promising for solving Ordinary Differential systems. However, in solving the PDE system, we are still looking for a robust model. In this paper, we presented a continuous CNN model that can continuously learn a PDE system. Traditional CNN models consider a continuous PDE system as a system with multiple layers, as shown in Fig.\ref{fig:tdnn}. It is a challenge for learning continuous time series in real-time.
Moreover, the sampling rate can be irregular wither higher frequency at the same time \cite{neil2016phased}.  As a result, despite the excellent performance of CNN in several research areas, this model fails to exhibit similar performance with continuous data.  One-dimensional CNN, aka a Time Delay Neural Network (TDNN) \cite{Lang88}, are used in some use cases such as Natural Language Processing (NLP). However, a continuous system such as an integral PDE system is a very complex for TDNN to achieve high accuracy. 

In this paper, we leverage the strength of a PDE solver and an ODE solver to define such a complex system. We proposed a new CNN model that acts like a one-dimensional TDNN to generate the integral of a PDE system and later convert that to an ODE system. Finally, an ODE solver is used to generate the output of the continuous system.  Here, a continuous system is a system that takes inputs continuously with higher frequency. For example, an IoT device was continuously collecting sensor data over time duration T. We have adopted the architecture of a dimensional Convolutional Neural Network similar to TDNN and introduced continuous learning by coupling a PDE and an ODE system in the model. We call it Continuous Convolutional Neural Network (CCNN).

The main contribution of this paper is as follows

\begin{itemize}
    \item A novel Continuous Convolutional Neural Network (CCNN).
    \item A new technique to learn PDE system using deep learning model.
    \item Finally, a comprehensive performance evaluation of the proposed model using a simulated system for Chirps. 
\end{itemize}

\section{Model Design}
\begin{figure}[htb]
    \centering
    \includegraphics[width=0.75\textwidth]{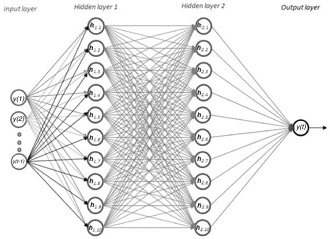}
    \caption{Time-delay neural network (TDNN) with two hidden layers}
    \label{fig:tdnn}
\end{figure}

The goal of this proposed model is to generalize Convolutional Neural Networks (CNNs) \cite{lecun1989backpropagation} to continuous space and time.  First, let us consider a TDNN with a single unit and generalize it to continuous time. Fig. \ref{fig:tdnn} shows the architecture of a TDNN model. 

In discrete time a TDNN model can be represented by Eq.~\eqref{eq:tdnn}
\begin{equation}
  y(t+1) = f( \sum_{\tau = 0}^T K(\tau, W_K, t) y(t-\tau) )
  \label{eq:tdnn}
\end{equation}
where $y(t)$ is the activity of the unit at time $t$, the function $f$ is a convenient nonlinearity, and $K(\tau, W_k)$ is a kernel function parameterized by $W_k$, typically using an explicit representation in the form of an array so $K(\tau, W_k) = W_k[\tau]$.  This is defined for $T_0 \leq t \leq T_1$, with boundary conditions (i.e., initialization) handled specially.

\begin{figure}
    \centering
    \includegraphics[width=0.8\textwidth]{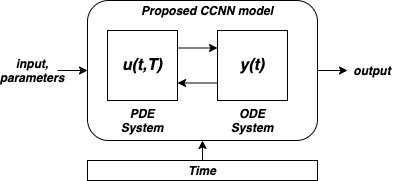}
    \caption{The block diagram of proposed Continuous CNN model}
    \label{fig:block}
\end{figure}

To move to continuous time, a TDNN model can be described as Eq.~\eqref{eq:tdnn-continous}.
\begin{equation}
  \ddt y(t) = f( \int_{\tau=0}^T K(\tau, W_k, t) y(t-\tau) \textrm{d}\tau )
  \label{eq:tdnn-continous}
\end{equation}
Here, the generalized kernel function $K(\tau, W_k, t)$ is converted as parametric to accept parameters. This is causal but involves an integral. (This can be viewed as a delay-differential equation.) When coding, this would mean that the differential form passed to the ODE solver would contain an invocation of an integration operator, to which the solution of $y(t')$ for $t'\leq t$ would be passed. However, typically, this is not available: only the current value of $y(t)$ is available. So instead, we encode the integral in a PDE as shown in Eq.~\eqref{eq:ut}(a).
\begin{subequations}
  \begin{gather}
    \ppx{\tau} u(t,\tau) = K(\tau, W_k, t) y(t-\tau) \\
    u(t,0) = 0
  \end{gather}
  \label{eq:ut}
\end{subequations}
with $0 \leq \tau \leq T$. Note that
\begin{equation}
  u(t,\tau') = \int_{\tau=0}^{\tau'} K(\tau, W_k, t) y(t-\tau) \textrm{d}\tau
  \label{eq:utau}
\end{equation}
So in particular for a time duration T=[0,T], Eq.~\eqref{eq:ut} provides the integral form of the PDE system.
\begin{equation}
  u(t,T) = \int_{\tau=0}^T K(\tau, W_k, t) y(t-\tau) \textrm{d}\tau
  \label{eq:uT}
\end{equation}
At this point, the integral from the driving term of $y$ can be removed, instead expressing it in terms of $u$ as shown in Eq.~\eqref{eq:uT}.
\begin{equation}
  \ddt y(t) = f(u(t,T))
  \label{eq:yt}
\end{equation}
Now we have a standard $PDE$ for $u$ as shown in Eq.~\eqref{eq:uT} and an ODE for $y$ as shown in Eq.~\eqref{eq:yt}, with no delays or integrals, and therefore Eq.~\eqref{eq:yt} system is suitable for standard solvers. Fig.\ref{fig:block} shows the block diagram of the CCNN model. In summary, CCNN is a coupled system of PDE and ODE systems.

\begin{figure*}[htb]
    \centering
    \includegraphics[width=\textwidth]{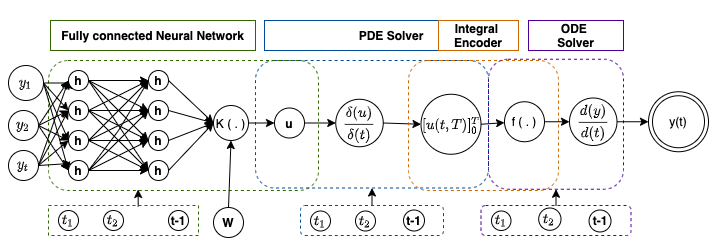}
    \caption{The architecture diagram of proposed Continuous CNN model}
    \label{fig:model}
\end{figure*}

Fig.~\ref{fig:model} shows the architecture diagram of proposed model. This continuous model consists of a fully connected TDNN model which takes input $Y=\{y_1, y_2, ... \ y_{t-1}\}$ continuously at each time step $T =\{t1, t2, ... \; t-1\}$. In addition, TDNN also takes the parameter $W_k$ for the kernel function $K$ as shown in Eq.~\eqref{eq:tdnn}. The second component in the model is a PDE solver which solves the input PDW system using Eq.~\eqref{eq:ut}(a). On the next step, the Integral encoder converts the PDE system in to its integral form $u(t,T)$. $u(t,T)$ is a suitable continuous equation and can be passed to an ODE solver similar to NODE models. Finally, the ODE solver in Fig.~\ref{fig:model} solves the continuous equation and generate the output $y_t$.
The proposed CCNN model can also handle high dimensional inputs. For example,  if $y(t) : \Re^n$ for $n \geq 2$, i.e., a vector, then the system immediately generalizes by adding some extra indices.

\section*{Implementation}
We have leveraged the neurodiffeq \cite{chen2020neurodiffeq} library to implement the proposed model. For the implementation, we have a PDE system described as Eq.~\eqref{eq:uT}, and an ODE system described as Eq.~\eqref{eq:yt}. Using TDNN neural network, we can have solutions for both of the systems.
\section*{Challenges}
To design the model, we need some additional sophisticated components for following challenges.  
\begin{itemize}
\item The $\ppx{\tau}u(t,\tau) = K(\tau, W_K, t) y(t-\tau)$ equation in Eq.\eqref{eq:ut}(a) has a shift, $t-\tau$, which might be hard to handle. Consider re-parameterizing $u$, with a change of variables, to get rid of the delay. Note that it would be okay to have a shift of the $t$ argument to $K$, since $K$ is fixed as far as the solver is concerned.

\item  A proper mechanism for driving term for an external input, instead of burying it in the initial conditions. Therefore, we have used the optimizer and model parameter as additional input to the model as shown in Fig.~\ref{fig:model}.

\item Need to design error function and adjoint system. In future work, we will implement adjoint ODE solver, for Now we have leveraged the ODE solver from Neurodiffeq \cite{chen2020neurodiffeq} library. 
\end{itemize}

\subsection*{Simulation with Chirps Signal}
\begin{figure}[htb]
    \centering
    \includegraphics[width=0.75\textwidth]{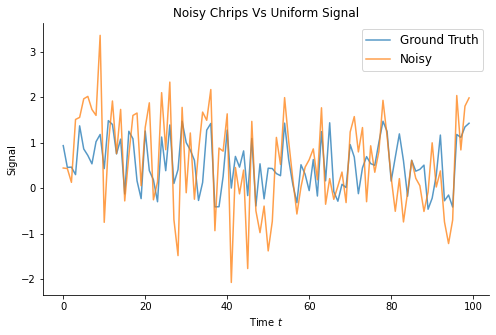}
    \caption{One-dimensional input containing “chirps” for proposed model}
    \label{fig:chrips}
\end{figure}

At this stage, we construct a one-dimensional input containing "chirps" to be detected against a background of white Gaussian noise, where the shape of a chirp varies systematically with time. This can be easily detected with a time-varying matched filter, and the performance evaluation shows that with learning, the kernel function $K$ converges to the correct time-varying matched filter. Fig.\ref{fig:chrips} shows the original signal in the blue curve and the noisy signal in the orange signal. 

The ODE system solver function can be described as algorithm~\ref{alg:ode}.

\begin{algorithm}
	\caption{ODE System Solver Function}
	\renewcommand{\algorithmicrequire}{\textbf{Input:}}
	\algorithmicrequire: input signal (y),time series (t), internal ($\tau$) and  weight ($W_k$) as well as bias parameters (b)
	\begin{algorithmic}[1]
		\Procedure{ode\_solver}{$y,t,parameters$}
		    \State \textbf{return}  $ODESOLVER(y,t)$
		\EndProcedure
	\end{algorithmic}
\label{alg:ode}
\end{algorithm}

Similarly, the PDE system solver function can be described as algorithm~\ref{alg:pde}. Here $kernel_function$ is the function shown in Eq.~\eqref{eq:uT}. The ODE solver function described in algorithm~\ref{alg:ode} is used in algorithm~\ref{alg:pde} as the PDE and ODE system are coupled. 

\begin{algorithm}
	\caption{PDE System Solver Function}
	\renewcommand{\algorithmicrequire}{\textbf{Input:}}
	\algorithmicrequire: input signal (y),time series (t), internal ($\tau$) and  weight ($W_k$) as well as bias parameters (b)
	\begin{algorithmic}[1]
		\Procedure{pde\_system\_solver\_function}{$y,t,parameters$}
		    \State$y\gets kernel\_function(y, W_k, b)$
		    \State$u,v \gets ode\_solver(y,t,parameters)$
		    \State \textbf{return}  $[diff(u, t, order=1) + diff(u, tau, order=1)]$
		\EndProcedure
	\end{algorithmic}
\label{alg:pde}
\end{algorithm}

The proposed model uses both ODE solver and PDE Solver modules. Appendix 5 shows the implementation of the proposed CCNN model.

\begin{figure}[htb]
    \centering
    \includegraphics[width=0.65\textwidth]{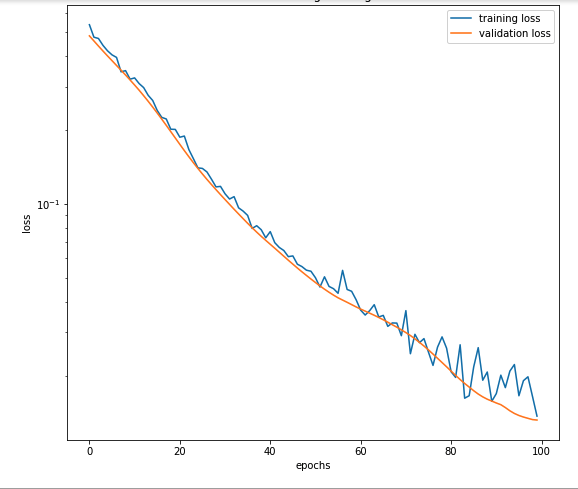}
    \caption{Training loss for proposed CCNN model One-dimensional input containing “chirps”}
    \label{fig:loss}
\end{figure}

Fig.~\ref{fig:loss} shows the training loss for the noisy signal in Fig.~\ref{fig:chrips}.

\section*{Conclusion}
This section proposed a new model for continuous Convolutional Neural Network (CCNN) and evaluated its performance for noisy signal (chips) processing. This model will help to process high dimensional time series with more than one dimension. This model also leverages the strength of Convolutional Neural Network (CNN) to process continuous data. In addition, any PDE system can be modelled and trained using this proposed model. 
\clearpage
\bibliographystyle{plain}
\bibliography{references}

\end{document}